%% file: neurips_2020.tex
\documentclass{article}

    \usepackage[preprint,nonatbib]{neurips_2020}

\usepackage[numbers]{natbib}

\usepackage[utf8]{inputenc} 
\usepackage[T1]{fontenc}    
\usepackage{hyperref}       
\usepackage{url}            
\usepackage{booktabs}       
\usepackage{amsfonts}       
\usepackage{nicefrac}       
\usepackage{microtype}      
\usepackage{epsfig}
\usepackage{graphicx}
\usepackage{amsmath}
\usepackage{amssymb}
\usepackage{subcaption}
\usepackage{caption}
\usepackage{algpseudocode}
\usepackage{enumitem}
\usepackage{hyperref}
\usepackage{times}
\usepackage{xcolor}
\usepackage{hyperref}
\usepackage{tikz}
\usepackage{physics}
\usepackage{mathdots}
\usepackage{yhmath}
\usepackage{cancel}
\usepackage{color}
\usepackage{siunitx}
\usepackage{array}
\usepackage{multirow}
\usepackage{gensymb}
\usepackage{wrapfig}
\usepackage{lipsum}
\usepackage{makecell}
\usepackage{nccmath}
\usepackage[export]{adjustbox}
\usepackage{tabularx}
\usetikzlibrary{fadings}
\usetikzlibrary{patterns}
\usetikzlibrary{shadows.blur}
\usetikzlibrary{shapes}

\newtheorem{prop}{Proposition}
\newcommand{\pdf}[1]{p_{#1}(#1)}

\DeclareMathOperator*{\argmax}{argmax}
\DeclareMathOperator*{\argmin}{argmin}
\title{Normalizing Flows Across Dimensions}

\author{%
  Edmond Cunningham\\
  University of Massachussets\\
  \texttt{edmondcunnin@cs.umass.edu} \\
   \And
   Renos Zabounidis \\
   University of Massachussets \\
   \texttt{rzabounidis@cs.umass.edu} \\   
   \And   
   Abhinav Agrawal \\
   University of Massachussets \\
   \texttt{aagrawal@cs.umass.edu} \\
   \And
   Ina Fiterau \\
   University of Massachussets \\
   \texttt{mfiterau@cs.umass.edu} \\
   \And
   Daniel Sheldon \\
   University of Massachussets \\
   \texttt{sheldon@cs.umass.edu} \\   
}

\begin{document}

\maketitle
\begin{abstract}

Real-world data with underlying structure, such as pictures of faces, are hypothesized to lie on a low-dimensional manifold.  This manifold hypothesis has motivated state-of-the-art generative algorithms that learn low-dimensional data representations.  Unfortunately, a popular generative model, normalizing flows, cannot take advantage of this.  Normalizing flows are based on successive variable transformations that are, by design, incapable of learning lower-dimensional representations. In this paper we introduce \textit{noisy injective flows} (NIF), a generalization of normalizing flows that can go across dimensions. NIF explicitly map the latent space to a learnable manifold in a high-dimensional data space using injective transformations. We further employ an additive noise model to account for deviations from the manifold and identify a \textit{stochastic inverse} of the generative process.  Empirically, we demonstrate that a simple application of our method to existing flow architectures can significantly improve sample quality and yield separable data embeddings.
\end{abstract}

\section{Introduction}
Normalizing flows \cite{rezende_variational_nodate,papamakarios_normalizing_2019} are a popular tool in probabilistic modeling, however they lack the ability to learn low-dimensional representations of the data and decouple noise from the representations. This could be a contributing factor to why normalizing flows lag behind other methods at generating high quality images \citep{kingma_glow_2018,ho_flow_2019,razavi_generating_2019,karras_analyzing_2020,song_generative_2019}.
The manifold hypothesis  \cite{fefferman_testing_2013} conjectures that real-world images, such as faces, lie on a low-dimensional manifold in a  high-dimensional space.  Consequently, one can expect that normalizing flows may not be able to properly represent data that satisfies the manifold hypothesis.  

The simplest method of obtaining a low-dimensional representation is by learning to map a lower dimensional vector to the data. The image of such a transformation will be a manifold in the data space \cite{ratli_multivariate_nodate}.  If the transformation is sufficiently expressive and the dimensionality of its domain matches that of the conjectured manifold, then the transformation may be able to learn the data manifold.  However if the transformation is bijective and the dimensionality of its domain is too large, it can at best learn a superset of the data manifold, and as a result map to points that are not data.  Normalizing flows use bijective functions that preserve dimension, so they are fundamentally incapable of perfectly modeling data that satisfies the manifold hypothesis.

Normalizing flows employ invertible functions to transform random variables \cite{rezende_variational_nodate}. It is the invertibility requirement that forces its input and output to have the same dimension. While this construction does not allow for low-dimensional representations, it affords exact log-likelihood  computation.  Log-likelihood-based inference is predicated on the ability to compute log-likelihood  \cite{casella_statistical_2002}, but this is rarely known exactly in deep machine learning models. For this reason, we would prefer to use low-dimensional representations to improve normalizing flows rather than seek a different method.

In this paper we introduce a generalization of normalizing flows which we call \textit{noisy injective flows}.  Noisy injective flows use injective functions to map across dimensions and a noise model to account for deviations from its learned manifold.  We show that this construction is a natural extension of normalizing flows that retains a form of invertibility while also decoupling its representation of data from extraneous noise.  We also provide an instance of noisy injective flows that can be incorporated into existing normalizing flow models to improve sample clarity without degrading log-likelihood  values.  Our contributions are summarized as follows:
\begin{itemize}
    \item We show that noisy injective flows are a generalization of normalizing flows that can learn a low-dimensional representation of data with a principled approach to account for deviations from the learned manifold.
    \item We introduce a stochastic inverse of the generative process for inference and training.
    \item We show that noisy injective flows have a simple mechanism to control how far samples deviate from the learned manifold. We showcase the flexibility of this mechanism when applied to image generation. A particular benefit of NIF is that we can vary the noise-model in a post-hoc manner to obtain crisper images and achieve higher metric based performance -- in terms of Fréchet Inception Distance \cite{heusel_gans_2018} and bits per dimension -- than normalizing flows.
\end{itemize}

\begin{figure}
    \centering
    \includegraphics[width=\linewidth]{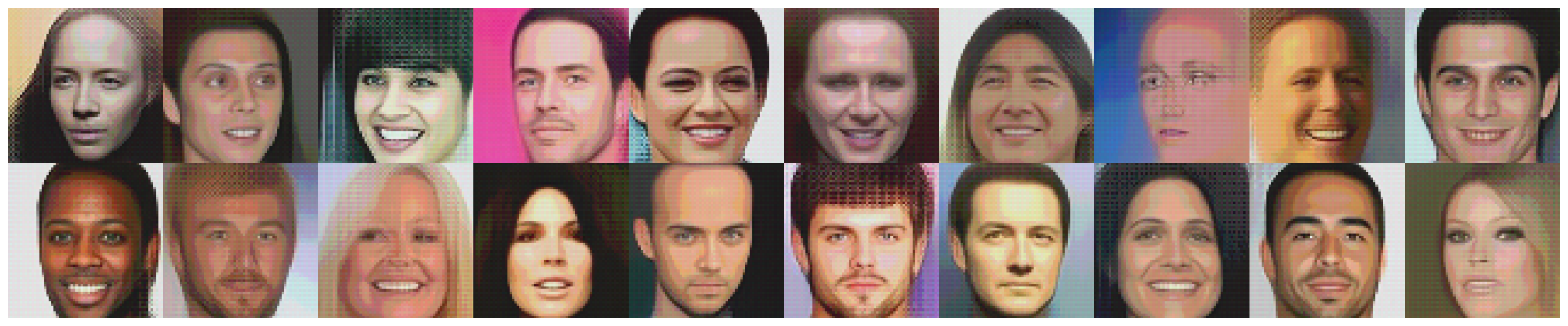}
    \caption{Generated faces from our method with a latent state size of 128.}
    \vspace{-0.25in}
    \label{fig:nif fig 1 sample}
\end{figure}

\section{Related Work}\label{section:related work}

The bulk of normalizing flows \cite{rezende_variational_nodate} research focuses on developing more powerful invertible layers \cite{ho_flow_2019}.  We, on the other hand, focus on improving the capabilities of normalizing flows to work across dimensions. \citet{gemici_normalizing_2016} were the first to apply normalizing flows across dimensions.  Their problem was constrained to when data was known to lie exactly on a manifold whose form in known analytically, but they did not investigate how to learn the manifold, nor how to treat data that is not on the manifold.  The recent work of \citet{brehmen_flows_2020} learns this manifold using a deterministic treatment of data that lies off the manifold and a term to penalize its distance from data, but does not provide a unified objective to perform maximum likelihood learning. \citet{kumar_regularized_2020} introduced a similar idea based on injective flows, using a novel lower bound on the injective change of variable formula for maximum likelihood training, however the authors note that their method does not work with data that does not lie exactly on the learned manifold.

Our work has similar features to variational autoencoders \cite{kingma_auto-encoding_2013} with Gaussian decoders.  The generative process we present can be seen as a special case of a variational autoencoder, but our use of injective functions, and our definition of a stochastic inverse makes our method resemble normalizing flows more closely.~\citet{dai_diagnosing_2019} consider the converse problem of ours -- how to use a method designed to model density around a manifold (VAEs with Gaussian decoders) for maximum likelihood learning, when data is exactly on a manifold.  We consider how to take an algorithm designed to learn density on a manifold (injective flows) for maximum likelihood learning when data lies around a manifold.  The algorithm they describe in their paper uses a 2-stage VAE that first learns the manifold and then learns an aggregate posterior that can be used for sampling whereas our model requires no such scheme. We do not compare against VAEs because we focus specifically on improving normalizing flows by incorporating low-dimensional representations.

\section{Noisy Injective Flows}\label{section:NIF}
Noisy injective flows are a generalization of normalizing flows that can be used to create normalizing flows across dimensions.  We start with a general change of variable formula as the foundation for our method and show that normalizing flows are derived as a special case.  Refer to section \ref{section:Gaussian NIF} for the form we use in experiments.

\subsection{Change of variable formula across dimensions}\label{subsection:change of var}
Let $z\sim p_z(z)$, $z\in \mathcal{Z}=\mathbb{R}^M$ and let $f_{\theta}:\mathcal{Z} \to \mathcal{X} \subseteq \mathbb{R}^{N}$ be an injective function parametrized by $\theta$. For $x' = f_{\theta}(z)$, the marginal distribution over $x'$ can be obtained using a generic change of variable equation \cite{noauthor_transforming_nodate}:
\begin{equation}\label{delta change of var}
    p_{x'}(x') = \int_{\mathbb{R}^M} p_z(z)\delta(x-f_{\theta}(z))dz
\end{equation}
When $N = M$, we can integrate over $z$ analytically to recover the well-recognized expression from normalizing flows~\cite{rezende_variational_nodate,papamakarios_normalizing_2019}:
\begin{align}
    p_{x'}(x')
    &=\int_{\mathbb{R}^N} \delta(x' - u)p_z(f_{\theta}^{-1}(u))\bigg|\frac{df_{\theta}^{-1}(u)}{du}\bigg|du \label{eq: sifting prop}\\
    &=p_z(f_{\theta}^{-1}(x'))\bigg|\frac{df_{\theta}^{-1}(x')}{dx'}\bigg| \label{eq: nf change of var}
\end{align}
But when the dimensionality of $x$ is greater than the the dimensionality of $z$, we can no longer analytically integrate because the integral in Eq.~\eqref{eq: sifting prop} will now be over $\mathcal{M}_{\theta}$ -- the manifold defined by the transformation $f_{\theta}$ :
\begin{align}
    p_{x'}(x')&= \underset{u \in \mathcal{M}_{\theta}}{\int} \delta(x'-u)p_z(f^{-1}_{\theta}(u))\bigg|\frac{df_{\theta}^{-1}(u)}{du}\frac{df_{\theta}^{-1}(u)}{du}^T\bigg|^{\frac{1}{2}}du \label{eq: manifold change of var}
\end{align}
This transformation changes dimensionality, so instead of a single Jacobian determinant we must use $|\frac{df_{\theta}^{-1}(u)}{du}\frac{df_{\theta}^{-1}(u)}{du}^T|^{\frac{1}{2}}$ to correctly relate the infinitesimal volumes $dz$ and $du$ \cite{noauthor_introduction_1975}.  However, the expression in Eq.~\ref{eq: manifold change of var} can be simplified for an $x'$ on $\mathcal{M}_{\theta}$. In particular, for $\forall x'\in \mathcal{M}_{\theta}$ we have the following injective change of variable formula \cite{gemici_normalizing_2016, kumar_regularized_2020}:
\begin{align}
    p_{x'}(x')= p_z(f^{-1}_{\theta}(x'))\bigg|\frac{df_{\theta}^{-1}(x')}{dx'}\frac{df_{\theta}^{-1}(x')}{dx'}^T\bigg|^{\frac{1}{2}} \label{injective change of var}
\end{align}
While this form gives us a normalizing flows like expression to evaluate, it may not be suitable for general probabilistic modeling; real data may not lie \emph{exactly} on a manifold but close to it. To account for such deviations, we propose an additive noise model.

\subsection{Adding noise to Injective Flows}\label{section:adding noise}
In Section~\ref{subsection:change of var}, we used $x'$ to denote the transformation of $z$. We define a new variable, $x$, as the sum of noise $\epsilon \sim \pdf{\epsilon}$ and $x'$: $x = x' + \epsilon$. As noise is assumed to be independent of $x'$, the density $p_{x}$ can be expressed using the convolution operator, denoted as *:
\begin{align}
    \pdf{x} &= \pdf{x'}*\pdf{\epsilon}
    = \int_{\mathbb{R}^M} p_z(z) p_\epsilon(x - f_{\theta}(z)) dz 
    \label{eq: conv change of var 1}
\end{align}
We note that there is a joint distribution in Eq.\eqref{eq: conv change of var 1} over \emph{latent} variable $z$ and \emph{observed} variable $x$, such that $p(x,z) = p_z(z) p_\epsilon(x - f_{\theta}(z))$\footnote{It is easy to check that the marginal condition is satisfied for $p(z)$: $\int p(x,z)dx=p_z(z)\int p_{\epsilon}(x-f_{\theta}(z))dx=p_z(z)\int p_{\epsilon}(\epsilon)d\epsilon=p(z)$}. For a given $z$, the accompanying generative story of $x$ is: evaluate $x' = f_{\theta}(z)$ and return $x=x' + \epsilon$ where $f_{\theta}$ is the parameterized injective function and $\epsilon \sim \pdf{\epsilon}$.
The introduction of $\pdf{\epsilon}$ renders our generative story non-deterministic.  Consequently, there is no deterministic method to invert $x$ -- we must instead construct a distribution $q(z|x)$ to map to the latent space.  In the spirit of normalizing flows, we choose $q(z|x)$ to be the \emph{stochastic inverse} of our generative story.

\subsection{Stochastic Inverse}
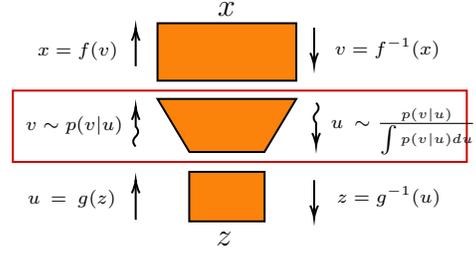
\begin{wrapfigure}{r}[-0.2in]{0.4\textwidth}
  \vspace{-6pt}
  \begin{center}
    \input{nif_schematic}
  \end{center}
  \caption{\small{Our method uses an stochastic invertible layer to build normalizing flows across dimensions.}}
  \label{fig:nif architecture}
  \vspace*{-0.2in}
\end{wrapfigure}
Noisy injective flows as discussed thus far are well specified generative models but lack a clear inference scheme.  We propose a specific choice for $q(z|x)$ to invert the generative process of $p_{\theta}(x|z)$: 
\begin{align}
    q_{\theta}(z|x)=\frac{p_{\theta}(x|z)}{\int p_{\theta}(x|z')dz'}\label{eq: normalized likelihood}
\end{align}
Firstly, note that this is not same as the posterior of the original model: Eq.~\eqref{eq: normalized likelihood} is the normalized likelihood distribution.  Alternatively, one can view this as the posterior distribution for an improper prior on $z$. Secondly, the above choice has nice properties for a Gaussian noise model: 

\begin{prop}\label{prop: pseudo inverse of NIF}
When $p_{\epsilon}(\epsilon)$ is a zero-centered Gaussian with covariance $\Sigma$, the modes of $q_{\theta}(z|x)$ are the solutions to $\underset{z}{\mathrm{argmin}}||x - f_{\theta}(z)||^2_{\Sigma^{-1}}$.
\end{prop}
To appreciate proposition~\ref{prop: pseudo inverse of NIF}, consider a data-point $x$ not on the manifold. One can expect the $z$ corresponding to the point on the manifold that is closest to $x$ to be a good representation for $x$. Our choice of $q_{\theta}(z|x)$ captures this intuition and places high probability mass on such points on the manifold.  

The main difference between the stochastic inverse and the posterior distribution is that the stochastic inverse does \emph{not} take into account the prior $p_z(z)$.  $q_{\theta}(z|x)$ infers $z$ solely based on how $p_{\theta}(x|z)$ generates $x$.  As a result, the stochastic inverse satisfies the analogy $p_{\theta}(x|z)$ is to $f_{\theta}(z)$ as $q_{\theta}(z|x)$ is to $f^{-1}_{\theta}(x)$.  In addition to extending the notion of an inverse for our generative process, $q_{\theta}(z|x)$ also affords an interpretable lower bound on the log-likelihood .

\subsection{Lower bounding log-likelihood }
Variational inference (VI)~\cite{jordan_introduction_1998} is a leading posterior approximation technique that use a parameterized distribution family $q_{\phi}$ to approximate the true posterior $p(z|x)$. In VI, one maximizes the lower bound to the marginal log-likelihood yielding an optimization problem equivalent to minimizing the Kullback–Leibler divergence from $q_{\phi}(z|x)$ to the true posterior. The following ELBO decomposition equation is central to the idea of VI:
\begin{align}
    \log p_x(x) &= \mathbb{E}_{q(z|x)}\bigg[\log\frac{p(x,z)}{q(z|x)}\bigg] + \mathit{KL}[q(z|x)||p(z|x)] \ge \underbrace{\mathbb{E}_{q(z|x)}\bigg[\log\frac{p(x,z)}{q(z|x)}\bigg]}_{\mathcal{L}}\label{eq: ELBO}
\end{align}
We use the ELBO to lower bound the log-likelihood, but do not learn $q_{\phi}$.  Instead, we use the stochastic inverse $q_{\theta}$ in place of the approximate posterior.  This choice simplifies the ELBO into two interpretable terms, one that defines log-likelihood  over $\mathcal{M}_{\theta}$ and one that will penalize a $\mathcal{M}_{\theta}$ that is far from data.  This choice new lower bound is specific to our model and notably cancels out the $\mathbb{E}_{q}[ \log {p(x|z)} ]$ term that appears in the standard ELBO decomposition.
{\small
\begin{align}
    \mathcal{L} & = \mathbb{E}_{q}\biggl[ \log {p_z(z)} \biggr ] + \log {\int p_{\theta}(x|z')dz'} \\
              & = \underbrace{\mathbb{E}_{q}\biggl[ \log {p_z(z)} \biggr ]}_{\text{Likelihood Term}} +  \underbrace{\log {\underset{x'\in \mathcal{M}_{\theta}}{\int} p_{\epsilon}(x - x')\bigg|\frac{df_{\theta}^{-1}(x')}{dx'}\frac{df_{\theta}^{-1}(x')}{dx'}^T\bigg|^{\frac{1}{2}} dx'}}_{\text{Manifold Term}} \label{simplified elbo}
\end{align}}
Related work on probabilistic models with manifolds consider log-likelihood  and separate term to capture distance from the manifold to data \cite{brehmen_flows_2020,kumar_regularized_2020}.  Our lower bound ends up using both of these terms in a statistically justified objective.  We note that the difference between $\log p_x(x)$ and $\mathcal{L}$ will always be nonzero because the construction of $q_{\theta}(z|x)$ yields $\mathit{KL}[q_{\theta}(z|x)||p(z|x)]>0$.  We do not find this to be problematic in practice and note that it is commonplace in VI to choose a model class for $q_{\phi}$ that does not include the true posterior, such as mean field VI \cite{hoffman_stochastic_nodate}.

\subsection{Implications on learning representations}
Noisy injective flows are constructed to decouple a low-dimensional representation of data from noise -- a capability that normalizing flows do not posses.  Normalizing flows use a single deterministic function to map between data and the latent space.  This transformation must learn every aspect of data at once.  As a result, they may not be able to decouple features of data from noise which can harm their sample quality.  Our method is constructed to not suffer from this problem and can instead learn the representation of data separately from extraneous noise.

\section{Gaussian Noisy Injective Flows}\label{section:Gaussian NIF}
We next give an instance of a noisy injective flow that is based on a Gaussian distribution.  We first describe the algorithm, then describe how it can be easily modified to scale to large images, incorporate non-linearities and yield a closed form log-likelihood .

We choose $p_{\epsilon}$ and $f_{\theta}$ so that we can sample from $p_{\theta}(x|z)$ and $q_{\theta}(z|x)$ efficiently and compute $\int p_{\theta}(x|z)dz$ in closed form:
{\small 
\begin{align}
    p_{\epsilon}(\epsilon)=N(\epsilon|b,\Sigma), \quad f_{\theta}(z)=Az, A\in \mathbb{R}^{M\times N}, M\leq N \label{gaussian nif}
\end{align}
}
Although this choice makes $\mathcal{M}_{\theta}$ a hyperplane, we can still create complex manifolds by transforming $x$ with a normalizing flow like in Fig.~\ref{fig:nif architecture}.  The mean, $Az$, can be used instead of a sample from $N(x|Az,\Sigma)$ in order to generate samples that lie on $\mathcal{M}_{\theta}$.  Below we give the closed form expressions of each quantity (we drop the dependence on $\theta$ for brevity.  See the appendix for a full derivation):
{\small\begin{align}
    p(x|z) = N(x|Az + b,\Sigma), \quad &q(z|x) = N(z|\Lambda^{-1} u,\Lambda^{-1}), \quad \log \int p(x|z)dz = \log Z_z - \log Z_x \label{model description},
\end{align}}
where 
{\small\begin{align}
    \mu &= x - b, \quad \Lambda = A^T \Sigma^{-1} A, \quad u = A^T \Sigma^{-1}\mu, \nonumber \\
    \log Z_z &= \frac{1}{2}(u^T \Lambda^{-1} u - \log |\Lambda| + \dim(z)\log(2\pi)), \nonumber \\
    \log Z_x &= \frac{1}{2}(\mu^T \Sigma^{-1} \mu + \log |\Sigma| + \dim(x)\log(2\pi)) \nonumber
\end{align}
}
To understand the role of $\log\int p(x|z)dz$ better, we make the simplifying assumption that $\Sigma=\sigma I$.
{\small 
\begin{align}
    \log\int p(x|z)dz = -\frac{1}{2\sigma}\mu^T(\mu - \overbrace{A^T(A^TA)^{-1}A\mu}^{\text{Projection of $\mu$ onto $\mathcal{M}_{\theta}$}}) - \frac{1}{2\sigma}\log|A^TA| - \frac{\dim(x)-\dim(z)}{2}\log(2\pi\sigma) \nonumber
\end{align}
}

We see that maximizing $\log\int p(x|z)dz$ will encourage the manifold to be close to data while accounting for the volume change of $z$.

Our algorithm has a time complexity of $O(\dim(z)^3)$ due to the calculation of the inverse and log determinant of $\Lambda$.  This is not an issue when $\dim(z)$ is small, but can become computationally prohibitive otherwise.  We next present a choice of $A$ that corresponds to efficient image upsampling.

\subsection{Nearest-neighbors up-sampling}
In general it is difficult to construct an $A$ that can be constructed using less than $O(\dim(z)\dim(x))$ space or yields a $\Lambda$ that can be inverted in better than $O(\dim(z)^3)$ time.  A situation where a naive implementation of Gaussian NIFs can become intractible is in generating high quality images.  Nearest-neighbor upsampling for progressive growing of images \cite{karras_progressive_2018} can alleviate this problem. Nearest-neighbor upsampling inserts a copy of each row and column in between an image's pixels.  This process can be written as a matrix vector product when we flatten the input image.  The resulting $\Lambda$ from equation~\ref{gaussian nif} is block diagonal and can therefore be inverted in $O(\dim(z))$ time.  As a result, the complexity of an NIF with Nearest-neighbor upsampling becomes $O(\dim(z))$.

\subsection{Stochastic coupling}
We can introduce non-linearities to Gaussian noisy injective flows using coupling transforms \cite{dinh_density_2017}.  Affine coupling is an invertible transformation that splits a vector $x$ into two components, $(x_1,x_2)$.  It sets $z_1=x_1$, uses non-linear functions $s$ and $t$ to get calculate $z_2=s(x_1)x_2+t(x_1)$ and then returns $z=(z_1,z_2)$.  The Jacobian determinant is equal to $\sum \log|s(x_1)|_i$.  

We can extend the notion of coupling to stochastic layers. Like in affine coupling, the input vector is split in two with one part unchanged.  However, we sample from a conditional distribution instead of computing a deterministic function: $z_1 = x_1$, $\;x_2\sim p_{\theta}(x_2|z_2;x_1)$ and $x_1 = z_1$, $\;z_2\sim q_{\theta}(z_2|x_2;x_1)$, and use the manifold term, $\log \int p_{\theta}(x_2|z_2;x_1)dz_2$, instead of the Jacobian determinant.

\subsection{Closed form log-likelihood }\label{analytical solution gaussian nif}
$p_x(x)$ can be computed analytically when $p_z(z)=N(z|0,I_m)$.  We reuse $u$, $\Lambda$ and $\log Z_x$ from Eq.~\eqref{model description} to get:
{\small\begin{align}
    p_x(x) &= \exp\{\log \hat{Z}_z-\log Z_x\}, \quad \text{where }\label{eq:closed form} \\
         \log \hat{Z}_z &= \frac{1}{2}(u^T (I_m+\Lambda)^{-1} u - \log |I_m+\Lambda| + \dim(z)\log(2\pi)) \nonumber
\end{align}}

This closed form solution yields a simple but powerful method to incorporate low-dimensional representations to normalizing flows.  The unit Gaussian prior that is used to train standard normalizing flows can be replaced with equation \eqref{eq:closed form} in order to gain give a normalizing flow the ability to learn a low-dimensional representation. We use this in our experiments to isolate the effect of using low-dimensional latent states.


\begin{figure}
    \centering
    \includegraphics[width=\textwidth]{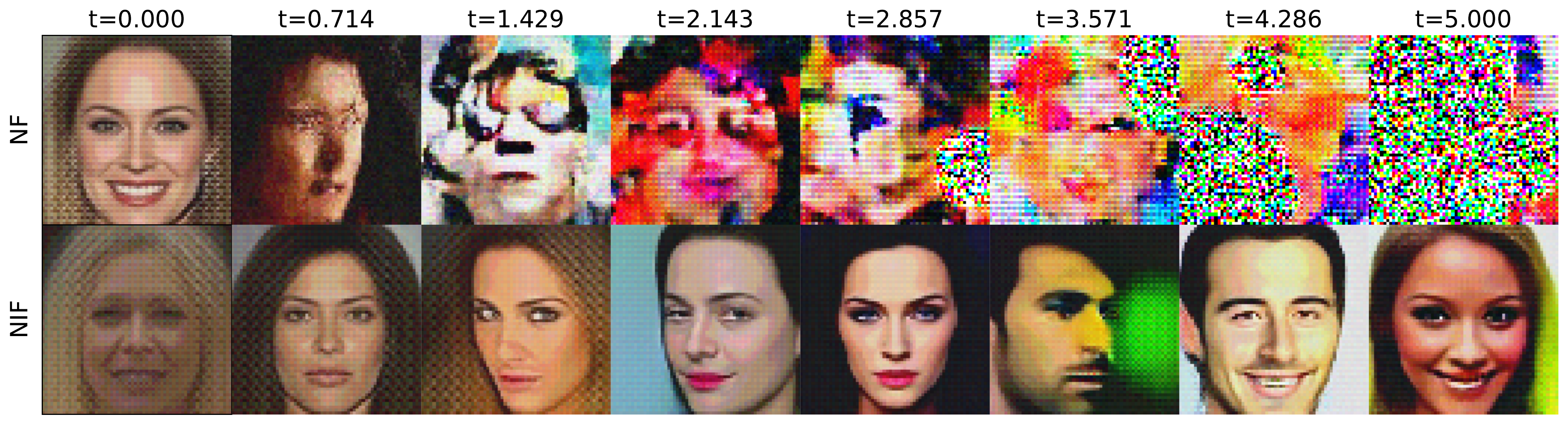}
    \caption{\small{Samples from priors with increasing variance (temperature). The top and bottom rows are standard normalizing flows and our method with a latent state size of 128 respectively.  Our method maps more of the latent space to the space of images than standard normalizing flows.}}
    \label{fig:temp comparison}
    \vspace*{-0.12in}
\end{figure}

\section{Experiments}\label{section:experiments}
The goal of our experiments is to demonstrate two main points: (1) low-dimensional latent states can significantly improve the learned representation of data over normalizing flows and (2) a single scalar value can be used control the sample quality of a trained NIF to ensure the NIF outperforms a comparable NF.  Our baseline normalizing flow uses a similar architecture to GLOW \cite{kingma_glow_2018} with 16 steps of their flow \cite{kingma_glow_2018}, each with 256 channels, and 5 multiscale components \cite{dinh_density_2017}. We define a comparable noisy injective flows to reuse the same normalizing architecture flow architecture, but replace the unit Gaussian prior with a closed form Gaussian NIF as described in section \ref{analytical solution gaussian nif}.  Our experiments use multiple latent state dimensionalities for the NIF.  We further isolate the effect of using a low-dimensional latent state by initializing and training the models with the batches of data by sharing random keys, which has the additional benefit of being fully reproducible. All of our code was written using the JAX \cite{jax2018github} Python library.
\subsection{Low dimensional representations}
 
Noisy injective flows bring the advantages of low-dimensional representations to normalizing flows.  We show that using low-dimensional latent states can help map more of its domain to faces, and also yield more separable data embeddings.

Both normalizing flows and noisy injective flows are trained to map a unit Gaussian in the latent space to data samples from the true data distribution.  However, one would expect that a good representation of data is able to generalize past a unit Gaussian and accordingly learn faces that are not from the dataset.  We employ temperature modeling \cite{kingma_glow_2018,chen_residual_nodate} to generate samples from more of the domain.  Temperature modeling achieves this by scaling the variance of the prior over $z$ by a scalar, $t$.  When $t=1.0$, we sample from the the original models.

\begin{wrapfigure}{r}[0cm]{0.7\linewidth}
\centering
\includegraphics[width=\linewidth]{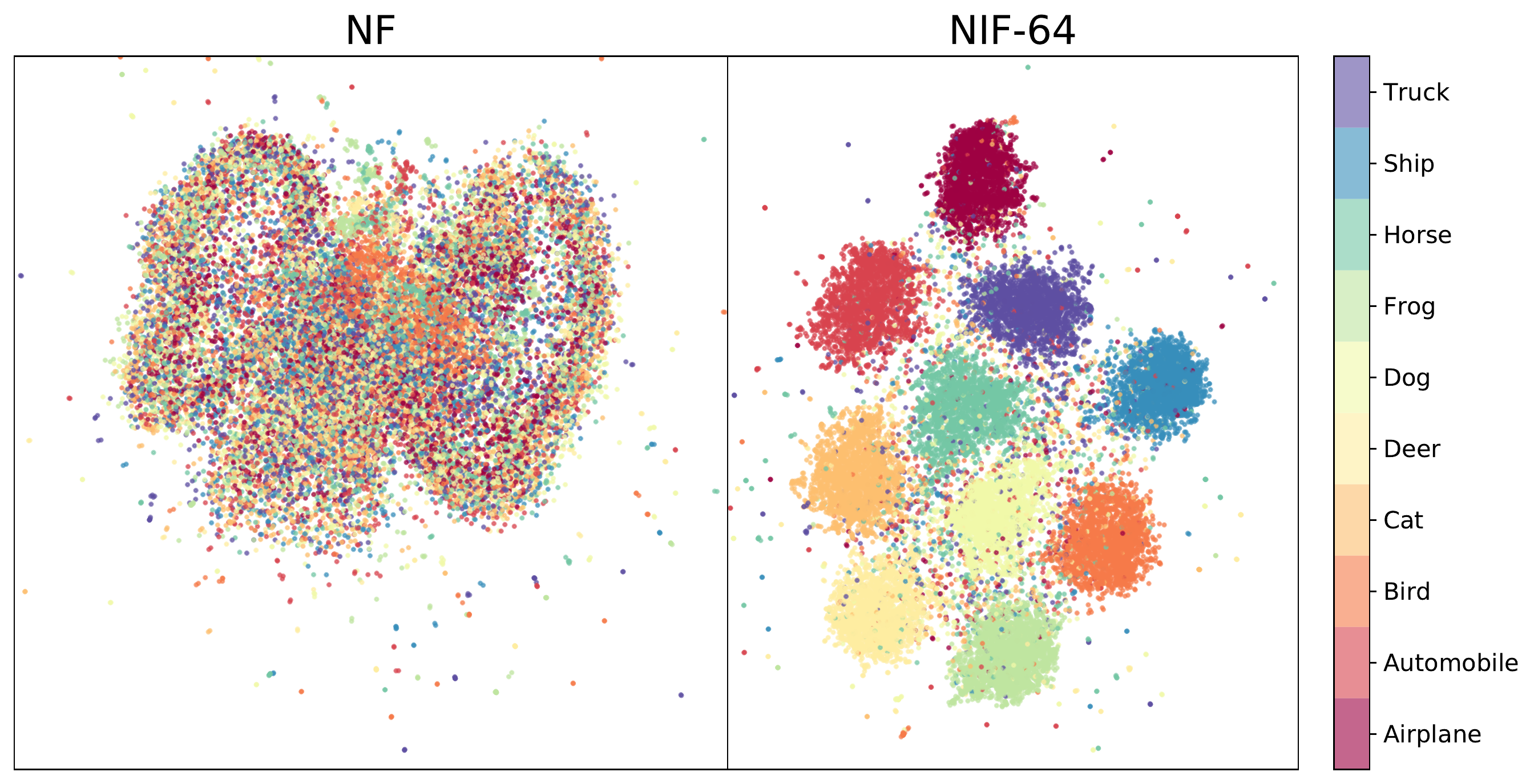}
\caption{\small{UMAP supervised embeddings of latent state of CIFAR test set.  Our method with a latent dimensionality of 64 on the right and a baseline normalizing flow on the left.}}
\label{fig:cifar}
\end{wrapfigure}

We see in Fig.~\ref{fig:temp comparison} that our method is able to generate faces for a large range of temperatures while normalizing flows can only generate faces for values of $t$ under $1.0$. 

Noisy injective flows also provide embeddings of the data that are more easily separable.  We use supervised UMAP \cite{mcinnes2018umap-software} to produce a low-dimensional embedding of the CIFAR-10 \cite{krizhevsky_learning_nodate} test set. Fig.~\ref{fig:cifar} shows that the NIF embedding cleanly separates the data from different classes while the NF embeddings cannot.

\begin{figure}
    \centering
    \includegraphics[width=\textwidth]{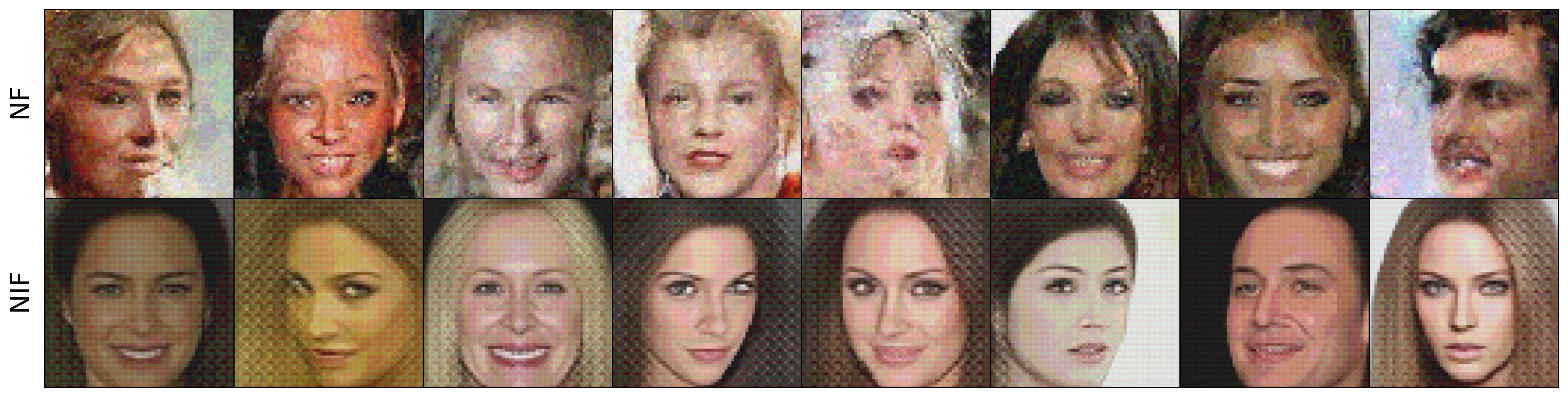}
    \caption{\small{Samples from a baseline normalizing flow (top) and a comparable noisy injective flow (bottom) with latent state dimensionality of 128. NIF can produce clear images by sampling directly on the learned manifold ($s=0.0$).}}
    \label{fig:nf vs nif vs if samples}
\end{figure}
\begin{figure}[b]
    \includegraphics[width=\linewidth]{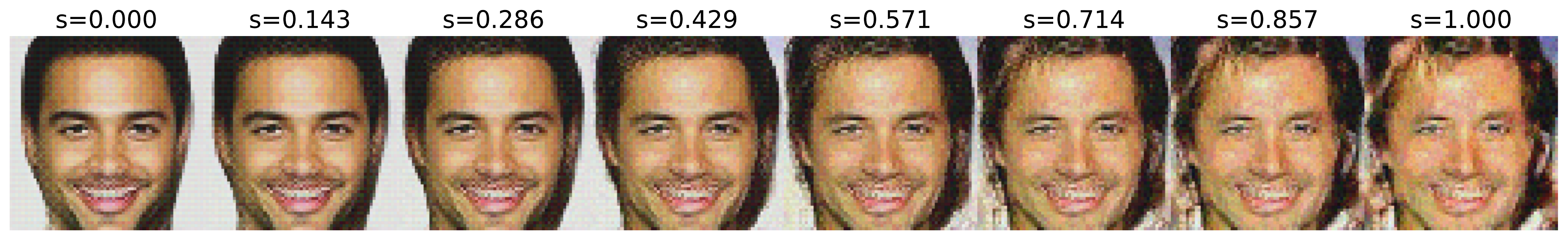}
    \caption{\small{Images from our model with the same latent state at varying distances from the manifold.  (Latent state dimensionality is 128)}}
    \label{fig:vary s}
\end{figure}
\subsection{Controlling deviations from the manifold}
We show that \emph{a single scalar parameter} introduced at test time can provide a simple method to control deviations from the manifold. The test time scalar parameter, $s$, controls the variance of a Gaussian NIF layer: $x\sim N(x|Az,s\Sigma)$. There are two notable settings of $s$: $s=1.0$ leaves the model unchanged while $s=0$ corresponds to the injective flow defined over the learned manifold $\mathcal{M}_{\theta}$.


Samples from our model when $s=0.0$ are generated directly on our learned manifold $\mathcal{M}_{\theta}$. In Fig.~\ref{fig:nf vs nif vs if samples}, we compare samples from the CelebA dataset \cite{liu2015faceattributes} from the baseline normalizing flow and from our method with $s=0.0$.  The samples generated on the manifold of the NIF are clearer and exhibit more cohesive facial structure than the samples from the normalizing flow. Samples from the manifold exhibit the high level features that our model has learned.  In the appendix we provide more samples from the manifold of models learned for Fashion MNIST and CIFAR-10.


We observe that increasing $s$ for a given point on the manifold will increase the amount of noise the sample exhibits. Fig.~\ref{fig:vary s} shows the effect of a sample as it is moved away from the manifold. Visually it may seem like deviating from the manifold serves no purpose other than adding random noise, however we find that small deviations from the manifold may add imperceptible features to the image. We find evidence of this in how the FID varies with $s$.
\begin{wrapfigure}{r}{0.45\textwidth}
    \centering
    \vspace{0mm}
    \begin{small}
    \begin{center}
    \includegraphics[width=\linewidth]{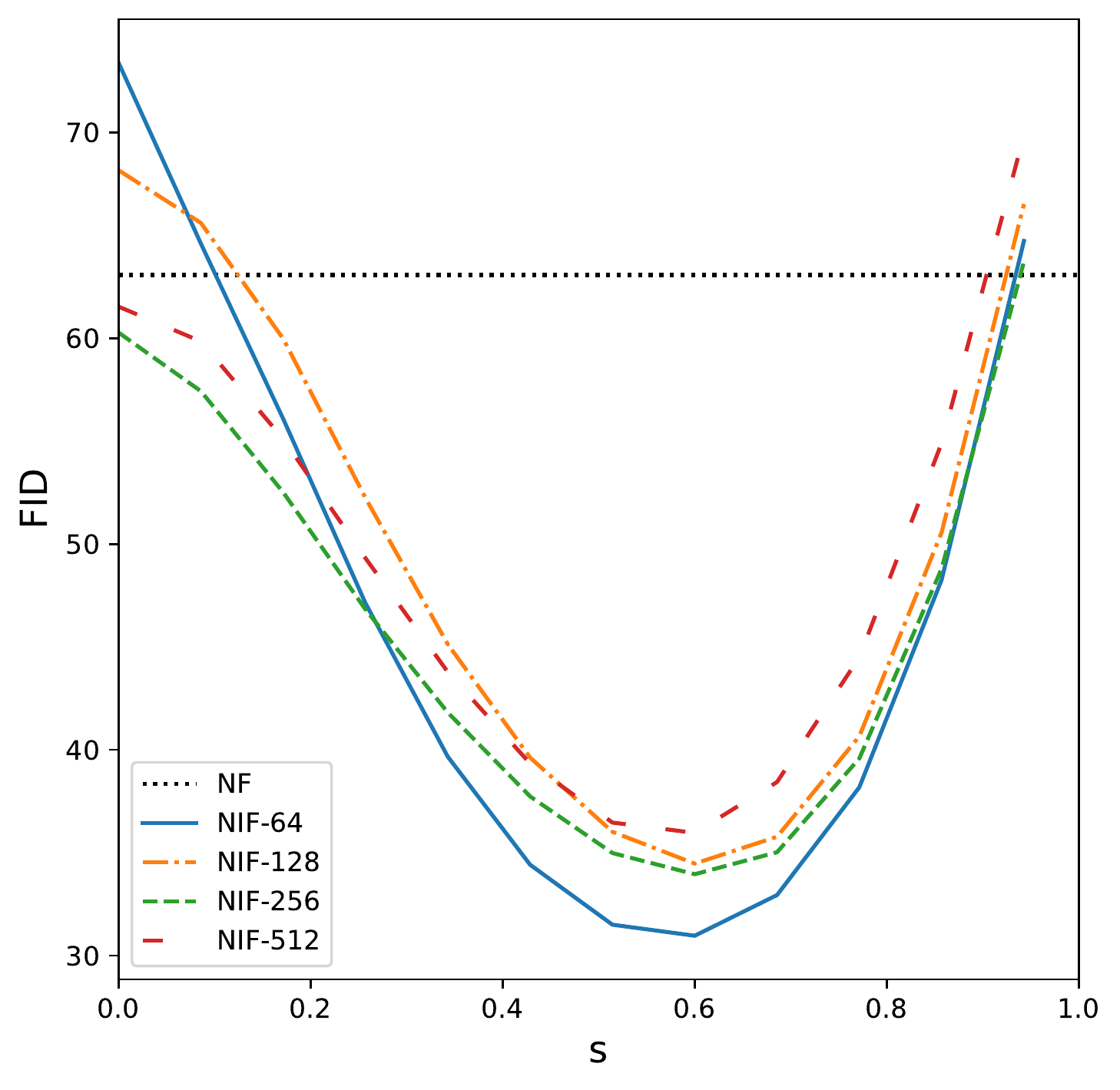}
    \caption{\small{FID with the CelebA dataset vs s.  Small deviations from the manifold provide significant improvements to FID. (Latent state dimensionality is 128)}}
  \label{subfig:celeba_fid_vary_s}
    \end{center}
    \end{small}
    \vspace{-4mm}
\end{wrapfigure}

Fréchet Inception Distance (FID) \cite{heusel_gans_2018} is a quantity used to measure the sample performance of a generative model. It computes a distance between two probability distributions by comparing the distributions of the activations of a state-of-the-art classifier for the Image-Net \cite{imagenet_cvpr09} dataset on samples from each dataset.  While FID has been shown to correlate with visual quality, at its core it can measure features that the classifier has learned. Fig.~\ref{subfig:celeba_fid_vary_s} shows that for some non-zero value of $s$, the resulting NIF can yield significantly better FID values. Given that non-zero values of $s$ do not correspond to clear visual changes in images, we can interpret the result of Fig.~\ref{subfig:celeba_fid_vary_s} to mean that slight deviations from the learned manifold correspond to noise in a \emph{feature space} that is perceptible to a classifier. By tuning $s$ over a random sub-sample of the training set and \emph{computing FID over a test set}, noisy injective flows are able to either match or significantly outperform normalizing flows in FID, as shown in Tab.~\ref{table:best fid scores}. In the appendix we investigate this phenomenon farther on the CIFAR-10 dataset~\cite{krizhevsky_learning_nodate}.


At $s=1.0$, we can evaluate if the latent dimensionality has a detrimental effect on log-likelihood. We see in table \ref{table:bits per dim} that this is not the case as noisy injective flows perform similar to or slightly better than normalizing flows in bits per dim across many latent state sizes and datasets.

\begin{table}%
  \centering
  \label{tbl:table}%
  \parbox{.49\linewidth}{
  \caption{\small{Fréchet Inception Distance (lower is better)}}
  \resizebox{\linewidth}{!}{
  \begin{tabular}{lccc}
    \toprule
    Model   & Fashion MNIST & CIFAR-10 & CelebA \\
    \midrule
    NF      & 42.77         & 78.58   & 63.07 \\
    NIF-64  & \textbf{23.97}         & 80.15   & \textbf{30.96} \\
    NIF-128 & 23.23         & 79.38   & 34.46 \\
    NIF-256 & 24.84         & 78.44   & 33.95 \\
    NIF-512 & 25.34         & \textbf{77.47}   & 35.96 \\
    \bottomrule
    \end{tabular}
    }
    \label{table:best fid scores}
    }
    \hfill
    \parbox{.49\linewidth}{
    \caption{\small{Bits per dimension (lower is better)}}
    \resizebox{\linewidth}{!}{
    \begin{tabular}{lccc}
    \toprule
     Model   & Fashion MNIST & CIFAR-10 & CelebA \\
     \midrule
     NF      &  1.518        & 1.072    & 0.852\\
     NIF-64  &  \textbf{1.506}        & \textbf{1.069}    & 0.839\\
     NIF-128 &  \textbf{1.506}        & 1.071    & 0.835\\
     NIF-256 &  1.515        & 1.073    & \textbf{0.830}\\
     NIF-512 &  1.523        & 1.070    & 0.838\\
     \bottomrule
    \end{tabular}
    }
    \label{table:bits per dim}
    }
\end{table}


\section{Conclusion}
We have presented a new probabilistic model, \textit{noisy injective flows}, that generalizes normalizing flows. The use of a stochastic inverse allows the method to transform across dimensions while maintaining the strengths of normalizing flows. We have presented an instance of NIFs that can be used to enhance existing flow models. We have demonstrated that our method was able to learn representations of data that are both low-dimensional and better than those learned by NFs. We also show that our model can be tuned to generate widely varied, high quality images, based on CelebA and Fashion MNIST. Noisy injective flows serve to bridge the gap between normalizing flows and state-of-the-art image generating methods while retaining the advantages of normalizing flows.



\bibliography{bibliography} 
\bibliographystyle{plainnat}

\newpage
\appendix

\section{Derivations}
\subsection{Notation}
\begin{align*}
    z &: \text{Latent variable in $\mathbb{R}^M$} \\
    \mathcal{Z} &: \text{Domain of z.  Equal to $\mathbb{R}^M$} \\
    p_z(z) &: \text{Prior over latent space} \\
    x &: \text{Ambient space random variable (data) in $\mathbb{R}^N$} \\
    \mathcal{X} &: \text{Domain of x} \\
    f_{\theta}(z) &: \text{Injective function that maps latent space to ambient (data) space, parametrized by $\theta$} \\
    \mathcal{M}_{\theta} &: \text{The manifold in $\mathbb{R}^N$ that is the image of $f_{\theta}(z)$} \\
    p_x'(x) &: \text{Probability density function over $\mathcal{M}_{\theta}$} \\
    p_x(x) &: \text{Probability density function over $\mathbb{R}^N$} \\
    p_\epsilon(\epsilon) &: \text{Noise model over $\mathcal{M}_{\theta}$} \\
    p_{\theta}(x|z) &: \text{Conditional likelihood of data given latent space.  Equal to $p_{\epsilon}(x-f_{\theta}(z))$} \\
    q_{\theta}(z|x) &: \text{Stochastic inverse of $p_{\theta}(x|z)$.  Equal to $\frac{p_{\theta}(x|z)}{\int p_{\theta}(x|z')dz'}$}
\end{align*}

\subsection{Equation \ref{delta change of var} - Change of variable formula}
\begin{align}
    p_x'(x') &= \frac{\partial }{\partial x'_1}\cdots\frac{\partial }{\partial x'_N}P(\mathcal{X}\leq x') \\
    &= \frac{\partial }{\partial x'_1}\cdots\frac{\partial }{\partial x'_N}P(f_{\theta}(\mathcal{Z})\leq x') \\
    &= \frac{\partial }{\partial x'_1}\cdots\frac{\partial }{\partial x'_N}\int_{\{z|f_{\theta}(z)\leq x'\}}p(z)dz \\
    &= \int_{\mathbb{R}^M}p(z)\frac{\partial }{\partial x'_1}\cdots\frac{\partial }{\partial x'_N}I[f_{\theta}(z)\leq x']dz \\
    &= \int_{\mathbb{R}^M}p(z)\delta(x'-f_{\theta}(z))dz
\end{align}

This general change of variable equation describes the probability density function of a transformed random variable.  When $f_{\theta)}$ is invertible and $M=N$, we can recover the standard normalizing flows change of variable formula.

\subsection{Equation \ref{eq: conv change of var 1} - Noisy injective flows marginal distribution}
\begin{align}
    p_x(x) = p_{x'}(x)*p_{\epsilon}(\epsilon) \\
           = \int p_{x'}(x-\epsilon)p_{\epsilon}(\epsilon)d\epsilon \\
           = \int \int p_{z}(z)\delta(x-\epsilon-f_{\theta}(z))dz p_{\epsilon}(\epsilon)d\epsilon \\
           = \int p_{z}(z)\int \delta(x-f_{\theta}(z)-\epsilon)p_{\epsilon}(\epsilon)d\epsilon dz  \\
           \text{we use the sifting property of the delta function to evaluate the integral} \nonumber \\
           = \int p_{z}(z) p_{\epsilon}(x-f_{\theta}(z)) dz
\end{align}
In section~\ref{section:adding noise} we showed that the convolved pdf is the marginal distribution over $x$ when the joint is defined as $p(x,z)=p_z(z)p_{\epsilon}(x-f_{\theta}(z))$.  However there is a more interpretable form of this equation that follows by letting $x'=f_{\theta}(z)$:
\begin{align}
    = \int_{\mathcal{M}_{\theta}}  p_{\epsilon}(x-x') p_{z}(f_{\theta}^{-1}(x'))|\frac{df^{-1}_{\theta}(x')}{dx'}\frac{df^{-1}_{\theta}(x')}{dx'}^T|^{\frac{1}{2}}dx'
\end{align}
This resulting equation has an intuitive explanation - the pdf of noisy injective flows is \emph{defined} as the expected value, over the noise model constrained to the learned manifold, of the injective change of variable formula from Eq.~\eqref{injective change of var}.  Although this form has no practical use, it serves to further justify the construction of noisy injective flows.

\subsection{Proposition \ref{prop: pseudo inverse of NIF} - Modes of the stochastic inverse are pseudo inverses}
The modes of $q(z|x)$ are at the values of $z$ that maximize $\log q(z|x)$.  If we assume that $p_{\epsilon}=N(\epsilon|0,\Sigma)$, we have:

\begin{align}
    \argmax_z \log q_{\theta}(z|x) \\
    = \argmax_z \log p_{\theta}(x|z) + \log\int p_{\theta}(x|z')dz' \\
    = \argmax_z \log p_{\theta}(x|z) \\
    = \argmax_z \log p_{\epsilon}(x-f_{\theta}(z)) \\
    = \argmax_z \log N(x-f_{\theta}(z)|0,\Sigma) \\
    = \argmax_z -\frac{1}{2}(x-f_{\theta}(z))^T\Sigma^{-1}(x-f_{\theta}) \\
    = \argmax_z -\frac{1}{2}||x-f_{\theta}(z)||_{\Sigma^{-1}}^2 \\
    = \argmin_z||x-f_{\theta}(z)||_{\Sigma^{-1}}^2
\end{align}

\subsection{Equation \ref{simplified elbo} - Evidence lower bound}

\begin{align}
    \mathcal{L} = \int q(z|x)\log\Big(\frac{p(x,z)}{q(z|x)} \Big)dz \\
     = \int q(z|x)\log\Big( \frac{p(x|z)p_z(z)}{\frac{p(x|z)}{\int p(x|z')dz'}} \Big)dz \\
     = \int q(z|x)\log\Big( p_z(z)\int p(x|z')dz' \Big)dz \\
     = \mathbb{E}_{q(z|x)}[\log p_z(z)] + \log \int p(x|z)dz
\end{align}

\subsection{Equation \ref{model description} - Gaussian NIF}

\begin{align}
    &N(x|Az+b,\Sigma) \nonumber\\
    &= \exp\{-\frac{1}{2}(x-Az-b)^T\Sigma^{-1}(x-Az-b) - \frac{1}{2}\log|\Sigma| - \frac{\dim(x)}{2}\log(2\pi)\} \\
    &= \exp\{-\frac{1}{2}(\underbrace{x-b}_{\mu}-Az)^T\Sigma^{-1}(x-b-Az) - \frac{1}{2}\log|\Sigma| - \frac{\dim(x)}{2}\log(2\pi)\} \\    
    &= \exp\{-\frac{1}{2}(\mu-Az)^T\Sigma^{-1}(\mu-Az) - \frac{1}{2}\log|\Sigma| - \frac{\dim(x)}{2}\log(2\pi)\} \\
    &= \exp\{-\frac{1}{2}z^TA^T\Sigma^{-1}Az + z^TA^T\Sigma^{-1}\mu - \underbrace{\frac{1}{2}[\mu^T\Sigma^{-1}\mu + \log|\Sigma| + \dim(x)\log(2\pi)]}_{\log Z_x}\} \\
    &= N^{-1}(z|A^T\Sigma^{-1}A,A^T\Sigma^{-1}\mu) \nonumber\\
    &\exp\{\frac{1}{2}[\mu^T\Sigma^{-1}A(\underbrace{A^T\Sigma^{-1}A}_{\Lambda})^{-1}\underbrace{A^T\Sigma^{-1}\mu}_{u} - \log|A^T\Sigma^{-1}A| + \dim(z)\log(2\pi)]\}\exp\{-\log Z_x\} \\
    &= N^{-1}(z|\Lambda,u)\exp\{\underbrace{\frac{1}{2}[u^T\Lambda^{-1}u - \log|\Lambda| + \dim(z)\log(2\pi)]}_{\log Z_z}\}\exp\{-\log Z_x\} \\
    &= N(z|\Lambda^{-1}u, \Lambda^{-1})\exp\{\log Z_z-\log Z_x\}
\end{align}

We use the names $\log Z_z$ and $\log Z_x$ because the values they represent are the log partition functions of $N(z|\Lambda^{-1}u, \Lambda^{-1})$ and $N(x|Az+b,\Sigma)$ respectively.

\subsection{Equation \ref{eq:closed form} - Closed form Gaussian NIF}

We start by proving the identity:
\begin{align}
    \int \exp\{-\frac{1}{2}z^TJz + z^Th\}dz = \exp\{\underbrace{\frac{1}{2}h^TJ^{-1}h - \frac{1}{2}\log|J| + \frac{\dim(z)}{2}\log(2\pi)}_{\log\hat{Z}_z}\}
\end{align}
Proof:
Consider a Gaussian probability density function: $N(z|J^{-1}h,J^{-1})$.  Because probability density functions integrate to 1, we have
\begin{align}
    \int N(z|J^{-1}h,J^{-1})dz = 1 \\
    \int \exp\{-\frac{1}{2}(z-J^{-1}h)^TJ(z-J^{-1}h) - \frac{1}{2}\log|J^{-1}| - \frac{\dim(z)}{2}\log(2\pi)\}dz = 1 \\
    \int \exp\{-\frac{1}{2}z^TJz + z^Th - \frac{1}{2}h^TJ^{-1}h + \frac{1}{2}\log|J| - \frac{\dim(z)}{2}\log(2\pi)\}dz = 1 \\
    \int \exp\{-\frac{1}{2}z^TJz + z^Th\}dz = \exp\{\frac{1}{2}h^TJ^{-1}h - \frac{1}{2}\log|J| + \frac{\dim(z)}{2}\log(2\pi)\}
\end{align}

With this identity, we can proceed with the main derivation:
\begin{align}
    p_x(x) = \int N(z|0,I_{m})N(x|Az + b,\Sigma) dz \\
    = \int \exp\{-\frac{1}{2}z^Tz - \frac{\dim(z)}{2}\log(2\pi)\}\\ \exp\{-\frac{1}{2}(x-Az-b)^T\Sigma^{-1}(x-Az-b) - \frac{1}{2}\log|\Sigma| - \frac{\dim(x)}{2}\log(2\pi)\} dz \\
    = \int \exp\{-\frac{1}{2}z^Tz -\frac{1}{2}z^T\underbrace{A^T\Sigma^{-1}A}_{\Lambda}z + z^T\underbrace{A^T\Sigma^{-1}(x-b)}_{u}\}dz \\
    \exp\{\underbrace{-\frac{1}{2}(x-b)^T\Sigma^{-1}(x-b) - \frac{1}{2}\log|\Sigma| - \frac{\dim(x)}{2}\log(2\pi)}_{-\log Z_x} - \frac{\dim(z)}{2}\log(2\pi)\} \\
    = \int \exp\{-\frac{1}{2}z^T(I_m + \Lambda)z + z^Tu\}dz \exp\{-\log Z_x - \frac{\dim(z)}{2}\log(2\pi)\} \\
    \text{We use the identity from above to introduce $\log\hat{Z}_z$}\nonumber \\
    = \int \exp\{-\frac{1}{2}z^T(I_m + \Lambda)z + z^Tu - \log \hat{Z}_z\}dz \exp\{\log \hat{Z}_z - \log Z_x\} \\
    = \int N(z|(I_m + \Lambda)^{-1}u,(I_m + \Lambda)^{-1})dz \exp\{\log \hat{Z}_z - \log Z_x\} \label{eq:closed form appendix} \\
    = \exp\{\log \hat{Z}_z - \log Z_x\} \label{eq:closed form appendix 2}
\end{align}

Eq.~\eqref{eq:closed form appendix 2} yields a simple equation that can be used to compute $p_x(x)$, however to embed an $x$ in the latent space, we use the pseudo-inverse of $x$ on the hyperplane, which takes the form:
\begin{align}
    z^+ = \Lambda^{-1}u
\end{align}
\newpage
\section{Additional Plots}

\subsection{Fashion MNIST}

\begin{figure}[h!]
    \centering
    \vspace{0mm}
    \begin{small}
    \begin{center}
    \includegraphics[width=\linewidth]{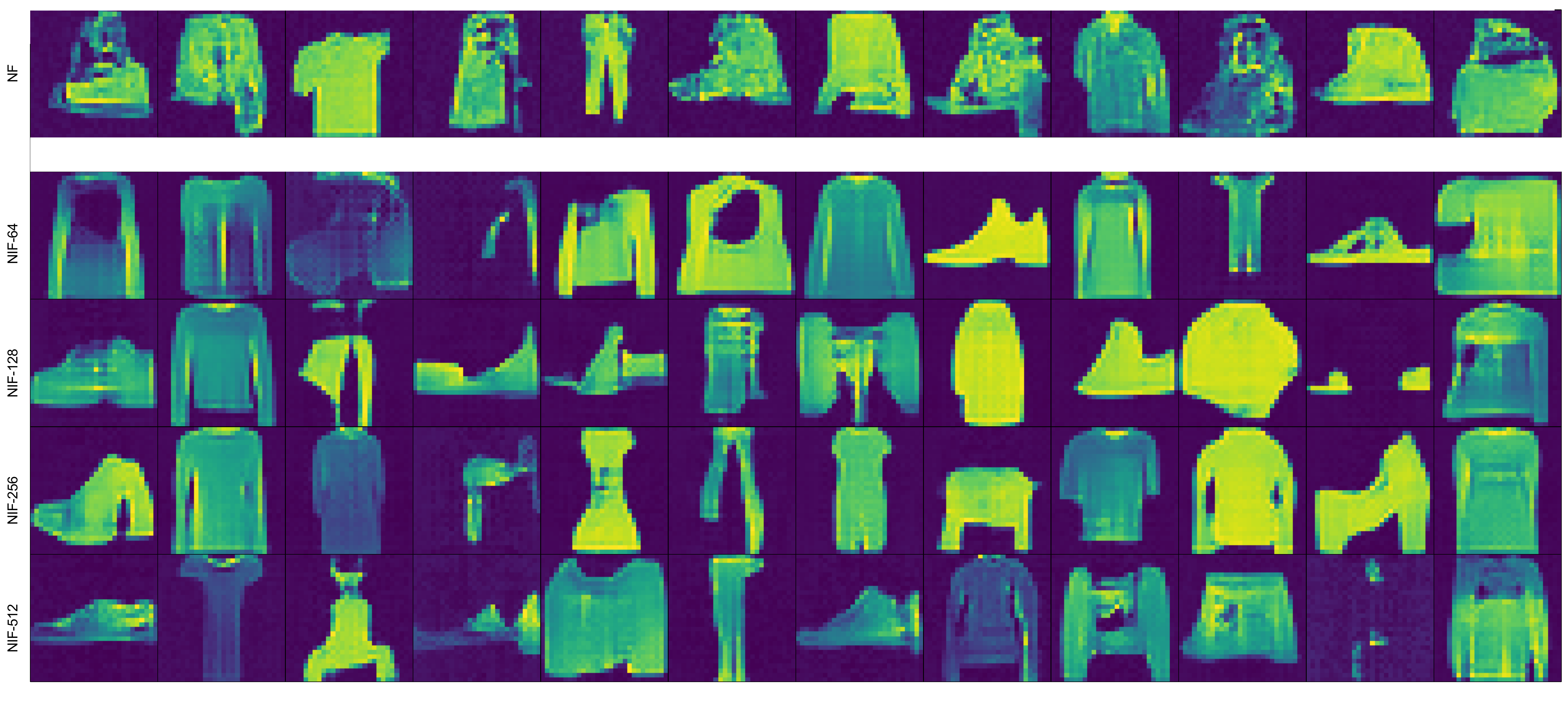}
    \caption{Samples from each model trained on Fashion MNIST.  Top row is from the baseline normalizing flow and, from top to bottom, the remaining rows are samples from a noisy injective flow with latent state dimensionalities of 64, 128, 256 and 512 respectively.  We see that even with small latent state dimensions, we are able to generate high }
    \label{fig:mnist samples}
    \end{center}
    \end{small}
    \vspace{-4mm}
\end{figure}

\subsection{CelebA Reconstructions}

\begin{figure}[h!]
    \centering
    \vspace{0mm}
    \begin{small}
    \begin{center}
    \includegraphics[width=\linewidth]{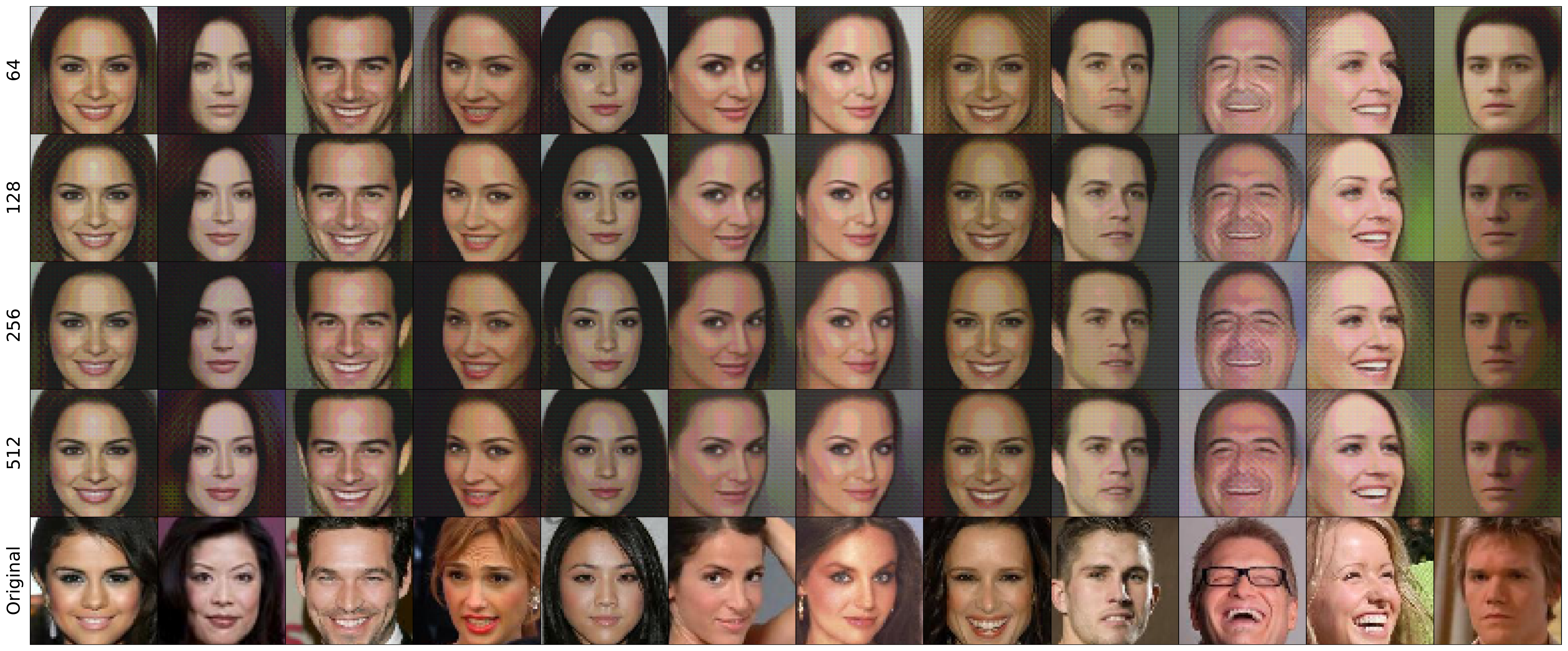}
    \caption{Reconstructions of CelebA samples from the manifold ($s=0.0$) of noisy injective flows with varying latent state sizes.  The rows, from top to bottom, use latent state sizes of 64, 128, 256 and 512.  The last row is the original image from the dataset.  We note that standard normalizing flows are constructed to give perfect reconstructions, so we omit them from this plot.}
    \label{fig:reconstructions}
    \end{center}
    \end{small}
    \vspace{-4mm}
\end{figure}
\newpage

\begin{figure}
    \begin{subfigure}{.3\textwidth}
        \includegraphics[width=\linewidth]{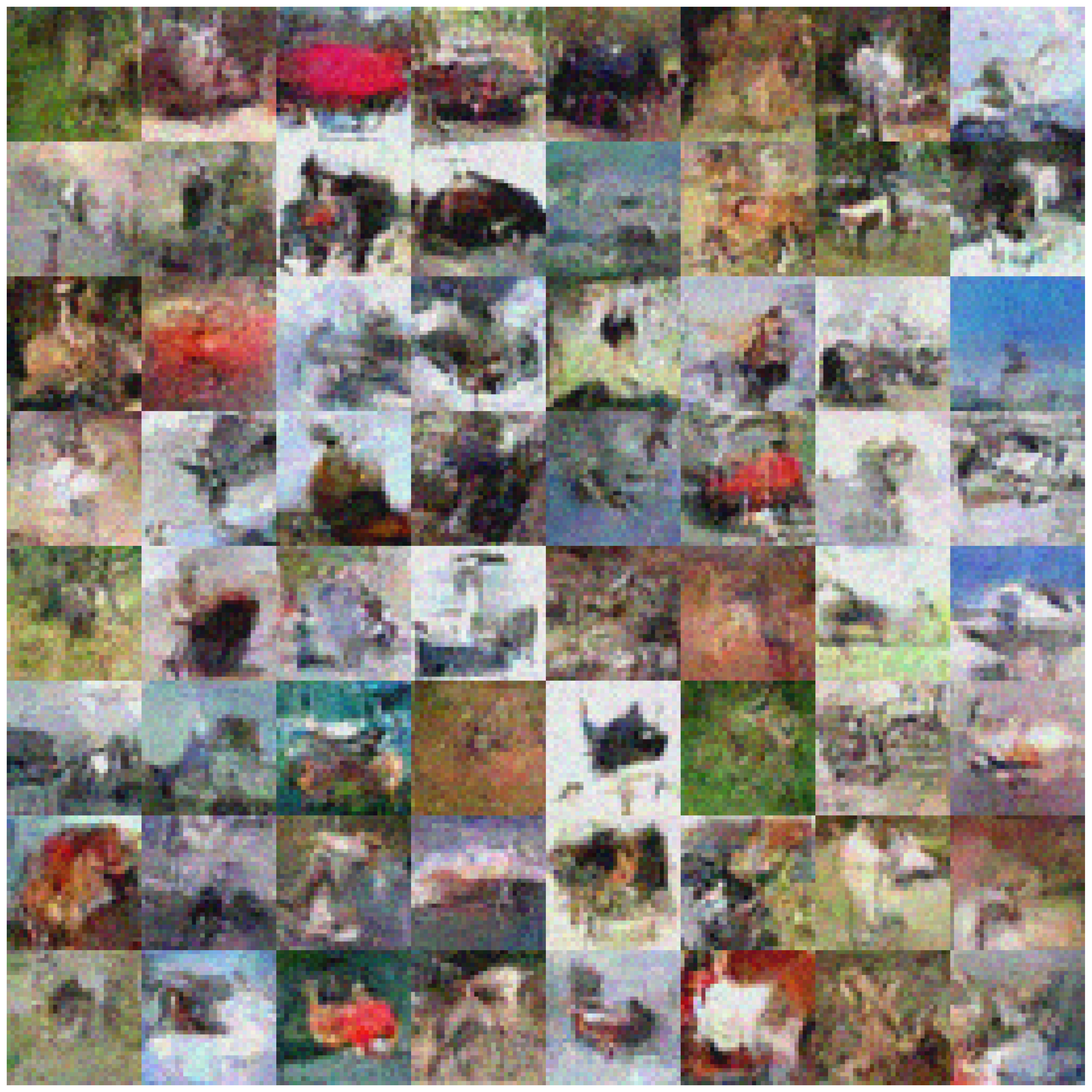}
        \caption{NF Samples.}
        \label{fig:cifar appendix nf}
    \end{subfigure}%
    \begin{subfigure}{.3\textwidth}
        \includegraphics[width=\linewidth]{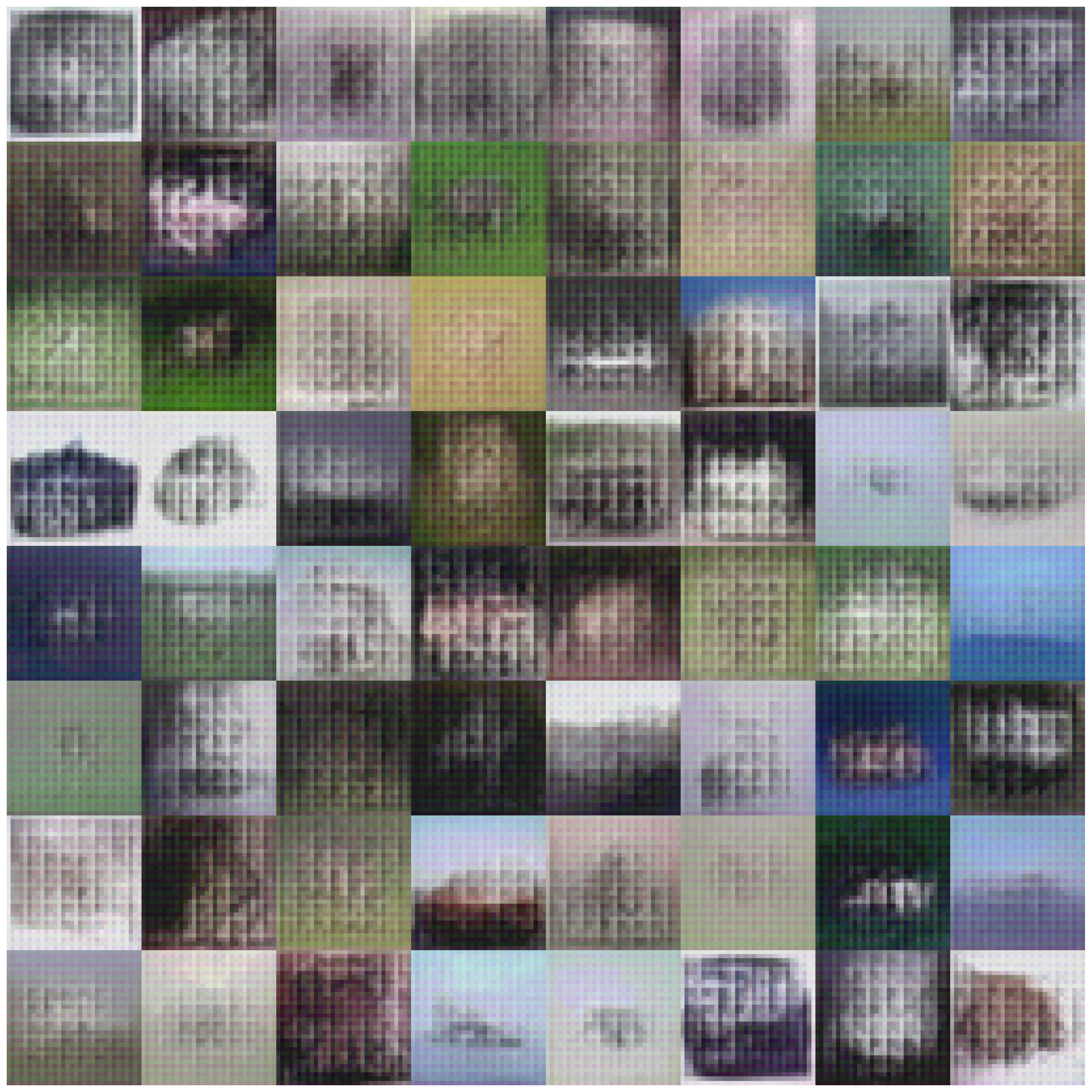}
        \caption{NIF samples (s=0.0).}
        \label{fig:cifar appendix nif s=0}    
    \end{subfigure}%
    \begin{subfigure}{.3\textwidth}
        \includegraphics[width=\linewidth]{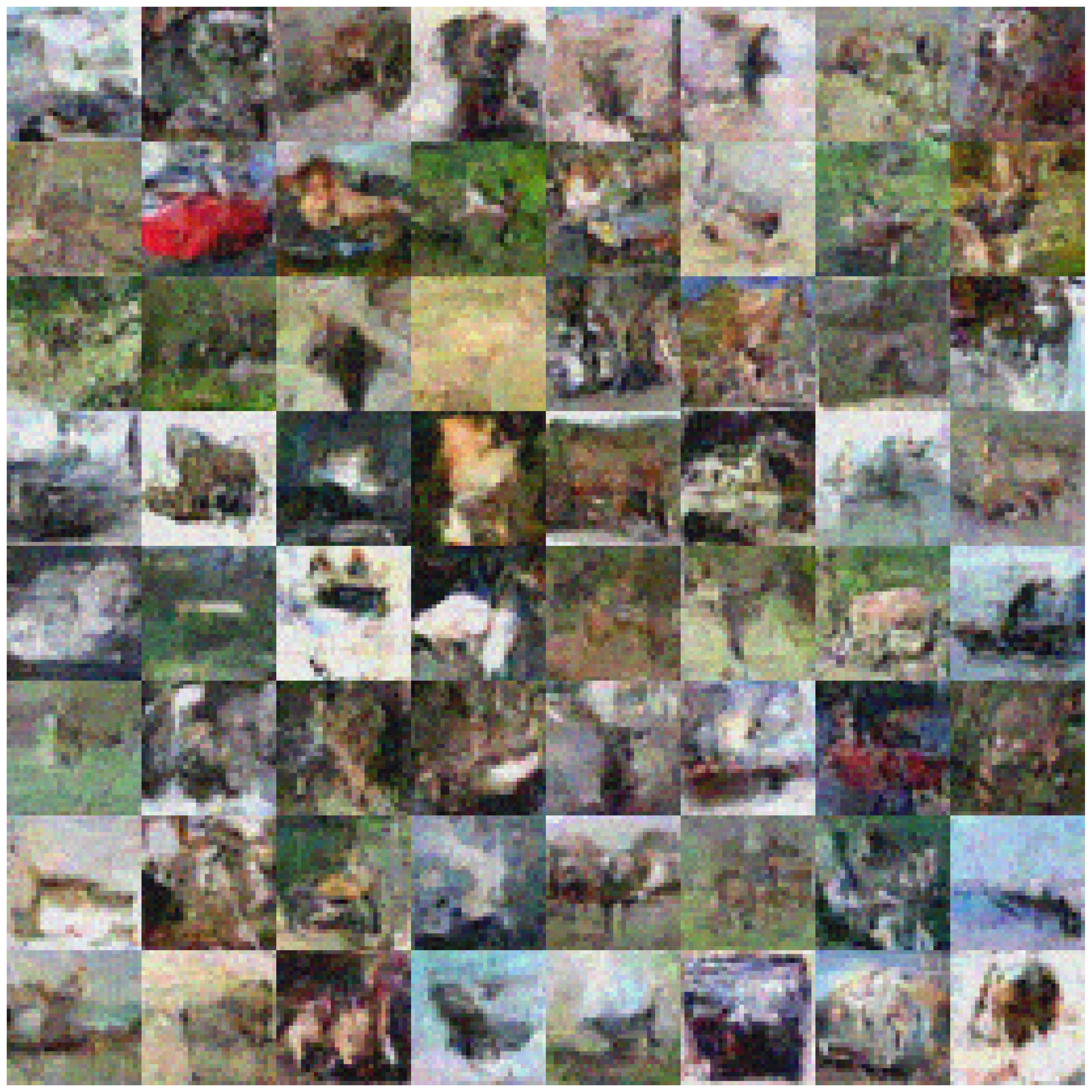}
        \caption{NIF samples (s=1.0).}
        \label{fig:cifar appendix nif s=1}    
    \end{subfigure}
    \caption{Samples from an NIF on its manifold can look worse than the samples from an NF, but will look similar away from the manifold.}
    \label{fig:cifar appendix}
\end{figure}

\newpage

\begin{wrapfigure}{r}{0.4\textwidth}
    \centering
    \vspace{-0.70in} 
    \begin{small}
    \begin{center}
    \includegraphics[width=\linewidth]{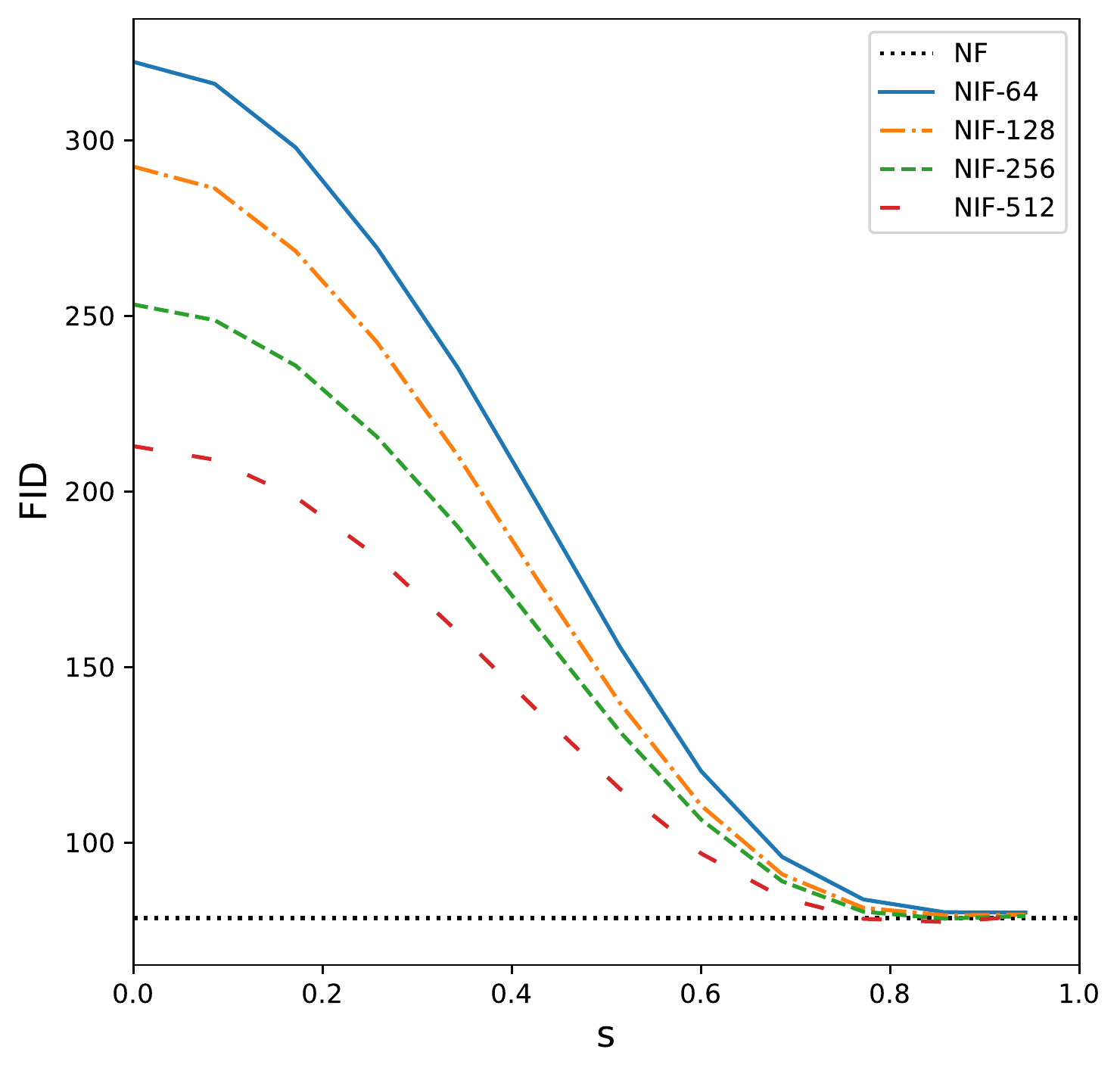}
    \caption{\small{FID vs s for the CIFAR-10 dataset.  The NIF models produce worse images than the NF close to the manifold, but approach the quality of the NF as $s$ approaches $1.0$.}}
  \label{subfig:appendix cifar vary s}
    \end{center}
    \end{small}
\end{wrapfigure}
\subsection{CIFAR-10 Results}
Noisy injective flows have a difficult time learning datasets that likely do not satisfy the manifold hypothesis such as CIFAR-10, however noisy injective flows can revert to the generative performance of normalizing flows by sampling off of the manifold.  Figure~\ref{fig:cifar appendix} shows samples from the baseline normalizing flow and noisy injective flow (with latent dimension of 128) from the experiments section.  The plot in the middle shows, figure~\ref{fig:cifar appendix nif s=0} samples from the manifold of the NIF.  The sample lack features of images that one expect to be present in CIFAR images.  However, when we sample from off the manifold ($s=1.0$) like in figure~\ref{fig:cifar appendix nif s=1}, noisy injective flows produce samples that resemble those from the normalizing flow.The plot of FID vs s in figure~\ref{subfig:appendix cifar vary s} provides a similar result.  The FID score of the NIF is poor when sampling on the manifold, but reverts back to that of the baseline normalizing flow as $s$ is increased to 1.
\newpage
\subsection{Deep noisy injective flow}
Here we show samples from a noisy injective flow whose architecture resembles figure~\ref{fig:nif architecture}.  This model used a latent state size of 128, used a low dimensional normalizing flow that consisted of 10 affine coupling layers, each with a 4 layer MLP with 1024 units in each hidden layer, and act norm and reverse layers in between each affine coupling.  A standard Gaussian NIF from section~\ref{section:Gaussian NIF} was used to change dimension into the same GLOW architecture described in the experiments section, but with 512 channels in each convolutional filter.

The use of a low dimensional normalizing flow allows the model to learn a probability density over the manifold.  Then the high dimensional flow is able to shape the manifold to fit data.  As a result, we see more variation in the images produced by this kind of noisy injective flow, especially at higher temperatures.

This model produced figure~\ref{fig:nif fig 1 sample}.  We found that it produced the best samples at higher temperatures ($t=1.5$ was used in figure~\ref{fig:nif fig 1 sample}).  Below we show samples from this model at varying temperatures.

\begin{figure}[h!]
    \centering
    \vspace{0mm}
    \begin{small}
    \begin{center}
    \includegraphics[width=\linewidth]{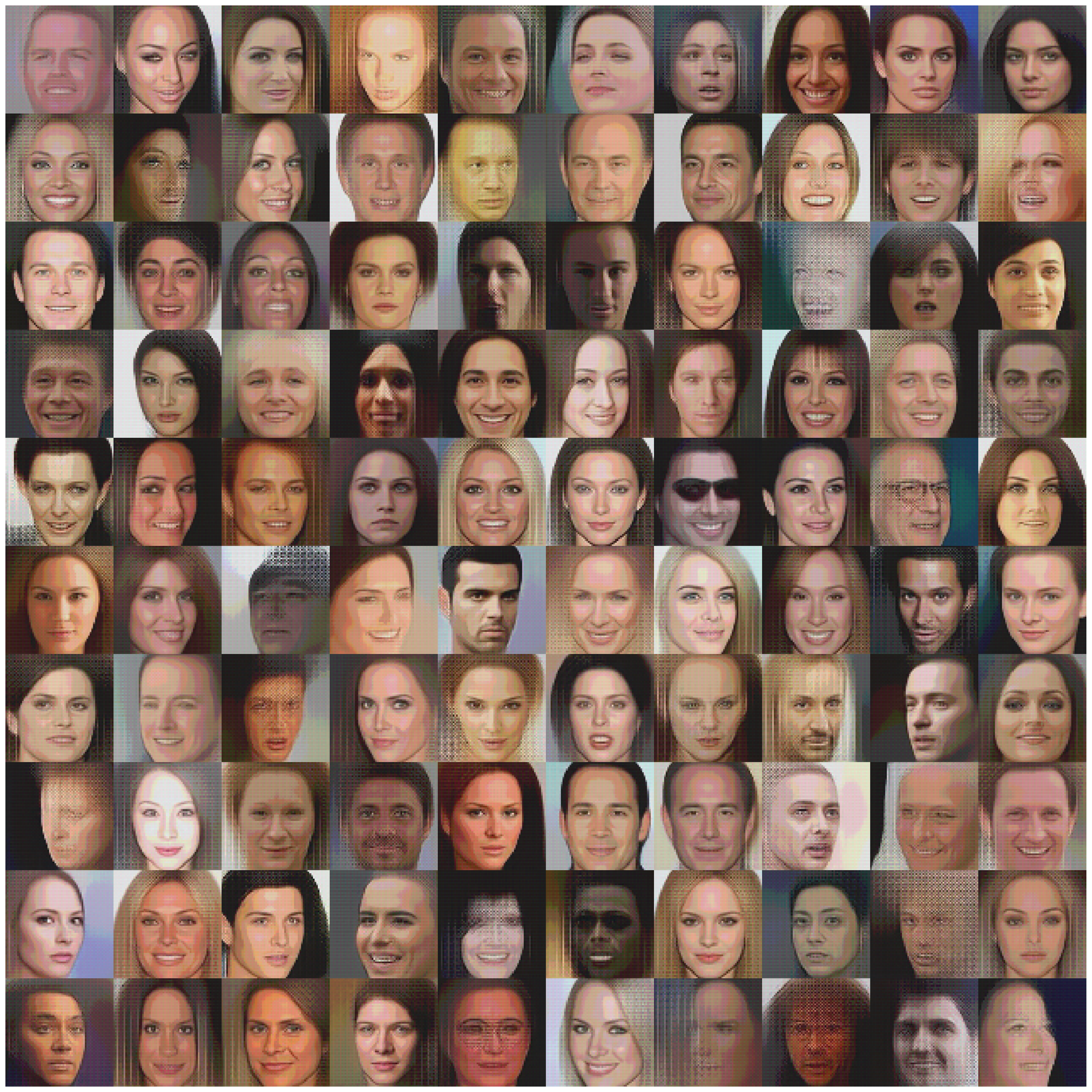}
    \caption{Samples from manifold of deep NIF at $t=1.0$}
    \label{fig:nif deep t=1}
    \end{center}
    \end{small}
    \vspace{-4mm}
\end{figure}

\begin{figure}[h!]
    \centering
    \vspace{0mm}
    \begin{small}
    \begin{center}
    \includegraphics[width=\linewidth]{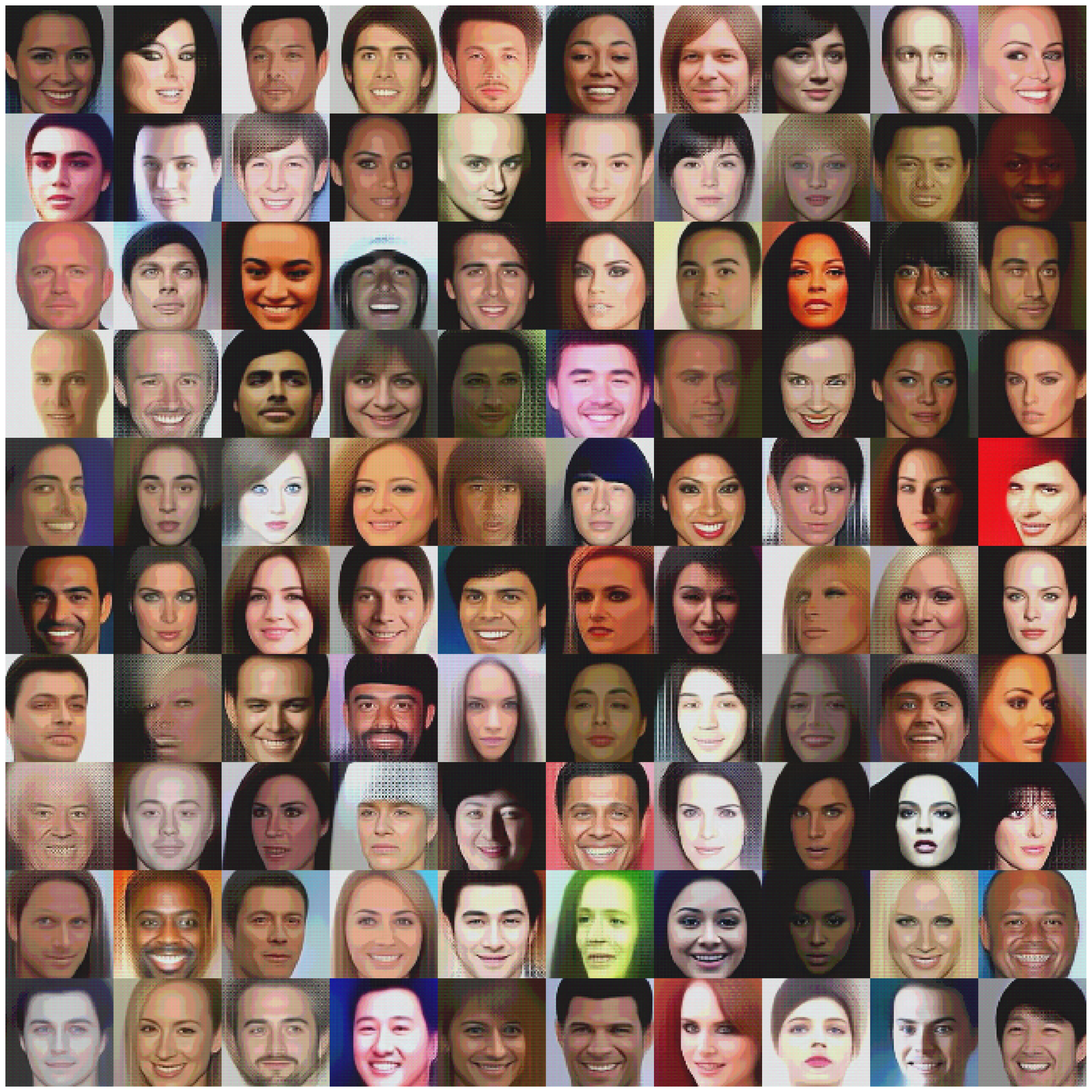}
    \caption{Samples from manifold of deep NIF at $t=2.0$}
    \label{fig:nif deep t=2}
    \end{center}
    \end{small}
    \vspace{-4mm}
\end{figure}

\begin{figure}[h!]
    \centering
    \vspace{0mm}
    \begin{small}
    \begin{center}
    \includegraphics[width=\linewidth]{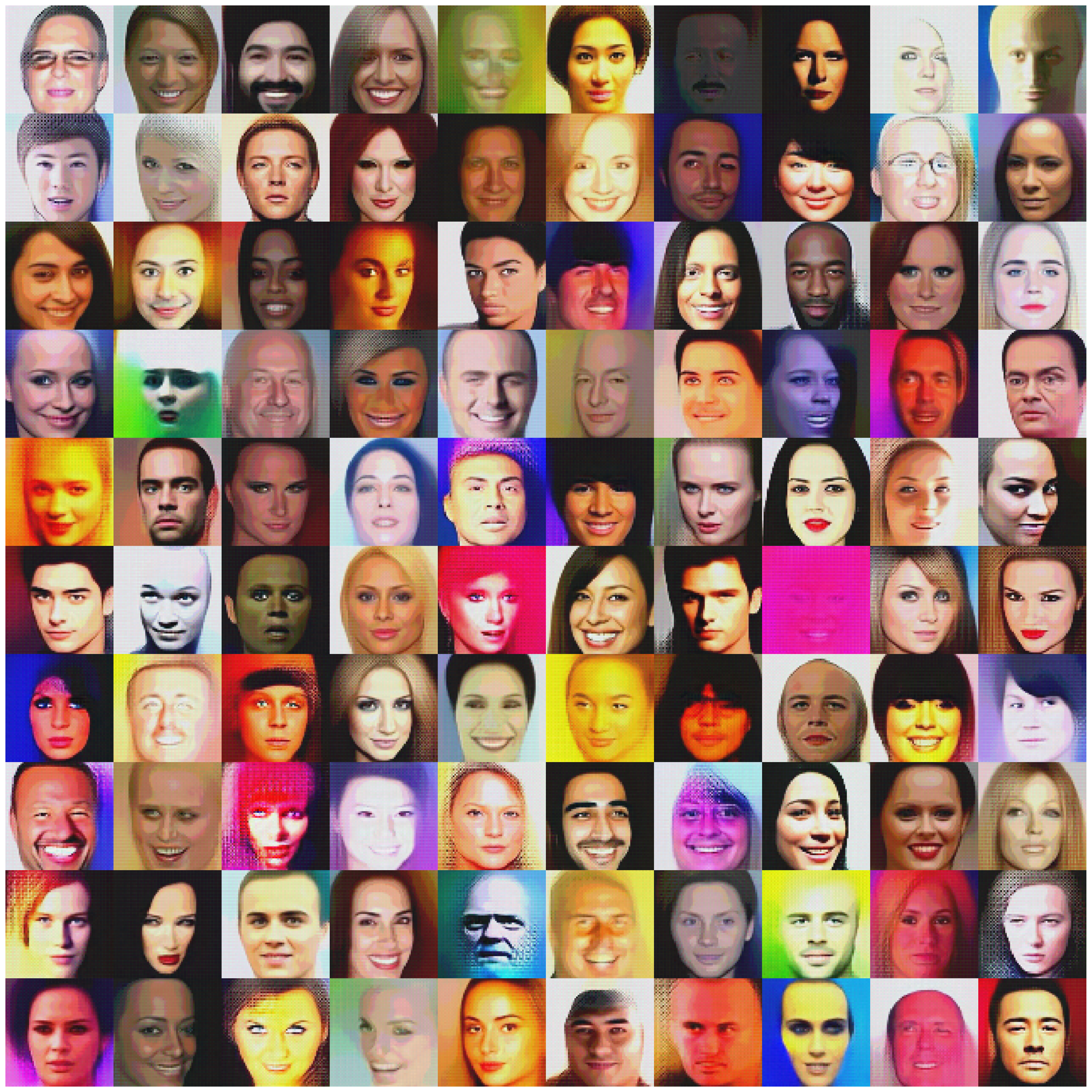}
    \caption{Samples from manifold of deep NIF at $t=4.0$}
    \label{fig:nif deep t=4}
    \end{center}
    \end{small}
    \vspace{-4mm}
\end{figure}

\end{document}

%% file: nif_schematic.tex
\tikzset{every picture/.style={line width=0.75pt}} 

\begin{tikzpicture}[x=0.75pt,y=0.75pt,yscale=-1,xscale=1]

\draw  [fill={rgb, 255:red, 251; green, 131; blue, 12 }  ,fill opacity=1 ] (281,37.8) -- (351,37.8) -- (351,66.8) -- (281,66.8) -- cycle ;
\draw  [fill={rgb, 255:red, 251; green, 131; blue, 12 }  ,fill opacity=1 ] (297,112.8) -- (335,112.8) -- (335,137.8) -- (297,137.8) -- cycle ;
\draw  [fill={rgb, 255:red, 251; green, 131; blue, 12 }  ,fill opacity=1 ] (351,76) -- (334.9,102.8) -- (297.1,102.8) -- (281,76) -- cycle ;
\draw    (270,59.51) -- (270,42) ;
\draw [shift={(270,40)}, rotate = 450] [color={rgb, 255:red, 0; green, 0; blue, 0 }  ][line width=0.75]    (6.56,-1.97) .. controls (4.17,-0.84) and (1.99,-0.18) .. (0,0) .. controls (1.99,0.18) and (4.17,0.84) .. (6.56,1.97)   ;
\draw    (360,40) -- (360,57.35) ;
\draw [shift={(360,59.35)}, rotate = 270] [color={rgb, 255:red, 0; green, 0; blue, 0 }  ][line width=0.75]    (6.56,-1.97) .. controls (4.17,-0.84) and (1.99,-0.18) .. (0,0) .. controls (1.99,0.18) and (4.17,0.84) .. (6.56,1.97)   ;
\draw    (270,100) .. controls (268.33,98.33) and (268.33,96.67) .. (270,95) .. controls (271.67,93.33) and (271.67,91.67) .. (270,90) -- (270,82) ;
\draw [shift={(270,80)}, rotate = 450] [color={rgb, 255:red, 0; green, 0; blue, 0 }  ][line width=0.75]    (6.56,-1.97) .. controls (4.17,-0.84) and (1.99,-0.18) .. (0,0) .. controls (1.99,0.18) and (4.17,0.84) .. (6.56,1.97)   ;
\draw    (270,137) -- (270,119) ;
\draw [shift={(270,117)}, rotate = 450] [color={rgb, 255:red, 0; green, 0; blue, 0 }  ][line width=0.75]    (6.56,-1.97) .. controls (4.17,-0.84) and (1.99,-0.18) .. (0,0) .. controls (1.99,0.18) and (4.17,0.84) .. (6.56,1.97)   ;
\draw    (360,117) -- (360,134.35) ;
\draw [shift={(360,136.35)}, rotate = 270] [color={rgb, 255:red, 0; green, 0; blue, 0 }  ][line width=0.75]    (6.56,-1.97) .. controls (4.17,-0.84) and (1.99,-0.18) .. (0,0) .. controls (1.99,0.18) and (4.17,0.84) .. (6.56,1.97)   ;
\draw  [color={rgb, 255:red, 208; green, 2; blue, 2 }  ,draw opacity=1 ] (208.05,71.8) -- (436.98,71.8) -- (436.98,107.8) -- (208.05,107.8) -- cycle ;
\draw    (361,78) .. controls (362.67,79.67) and (362.67,81.33) .. (361,83) .. controls (359.33,84.67) and (359.33,86.33) .. (361,88) -- (361,89.8) -- (361,97.8) ;
\draw [shift={(361,99.8)}, rotate = 270] [color={rgb, 255:red, 0; green, 0; blue, 0 }  ][line width=0.75]    (6.56,-1.97) .. controls (4.17,-0.84) and (1.99,-0.18) .. (0,0) .. controls (1.99,0.18) and (4.17,0.84) .. (6.56,1.97)   ;

\draw (219,45.4) node [anchor=north west][inner sep=0.75pt]  [font=\scriptsize]  {$x=f( v)$};
\draw (214,120.4) node [anchor=north west][inner sep=0.75pt]  [font=\scriptsize]  {$u\ =\ g( z)$};
\draw (213,83.4) node [anchor=north west][inner sep=0.75pt]  [font=\scriptsize]  {$v\sim p( v|u)$};
\draw (367,79.4) node [anchor=north west][inner sep=0.75pt]  [font=\scriptsize]  {$u\ \sim \frac{p( v|u)}{\int p( v|u) du}$};
\draw (369,43.4) node [anchor=north west][inner sep=0.75pt]  [font=\scriptsize]  {$v=f^{-1}( x)$};
\draw (370,118.4) node [anchor=north west][inner sep=0.75pt]  [font=\scriptsize]  {$z=g^{-1}( u)$};
\draw (309,141.4) node [anchor=north west][inner sep=0.75pt]  [font=\large]  {$z$};
\draw (310,25.4) node [anchor=north west][inner sep=0.75pt]  [font=\large]  {$x$};

\end{tikzpicture}